\documentclass[conference]{IEEEtran}
\usepackage{amsmath,amssymb}
\usepackage{graphicx}
\usepackage{booktabs}
\usepackage{multirow}
\usepackage{xcolor}
\usepackage{tikz}
\usetikzlibrary{arrows.meta,positioning,calc}
\usepackage{pgfplots}
\pgfplotsset{compat=1.16}
\usepgfplotslibrary{groupplots}
\usepackage[hidelinks]{hyperref}
\usepackage[nocompress]{cite}
\usepackage{algorithm}
\usepackage{algpseudocode}

\newcommand{\qh}{\hat{q}}
\newcommand{\lh}{\hat{\lambda}}
\newcommand{\ph}{\hat{p}}

\begin{document}

\title{ReCo: When to Relocate Sensor Kits under Deployment Constraints---A NILM Case Study}

\author{\IEEEauthorblockN{Haokun Chen}
\IEEEauthorblockA{Independent Researcher\\
Shanghai, China\\
haokunchen1217@gmail.com}
\and
\IEEEauthorblockN{Yu Tong}
\IEEEauthorblockA{Shanghai Eneintel Technology Co., Ltd.\\
Shanghai, China\\
tongyu@eneintel.com}
\and
\IEEEauthorblockN{Yehai Chen}
\IEEEauthorblockA{Shanghai Eneintel Technology Co., Ltd.\\
Shanghai, China\\
chenyehai@eneintel.com}}

\IEEEoverridecommandlockouts
\IEEEaftertitletext{\noindent\begin{minipage}{\textwidth}
\footnotesize
\textcopyright{} 2026 IEEE. Personal use of this material is permitted. Permission from IEEE must be obtained for all other uses, in any current or future media, including reprinting/republishing this material for advertising or promotional purposes, creating new collective works, for resale or redistribution to servers or lists, or reuse of any copyrighted component of this work in other works.
\end{minipage}\par\vspace{\baselineskip}}
\maketitle

\begin{abstract}
Many sensing tasks obtain training labels only by deploying instruments in the field. With a limited number of sensor kits, a collection deadline, and measurement downtime at every move, the collector must repeatedly decide whether to stay at the current site or relocate. We study this decision in non-intrusive load monitoring (NILM), which estimates the power drawn by individual appliances from a home's main meter and is trained on data from homes temporarily fitted with appliance-level sub-meters. In NILM, appliance usage varies with the appliance, season and climate, and the value of new data depends on how diverse the combinations of target operation and background load are. To address this, we propose a constraint-based relocation framework and instantiate it for NILM as ReCo (Relocation by Coverage gain). ReCo counts new operating regimes in a joint target--background feature space, forecasts each home's future gain from the data collected so far, and each night weighs the gain of staying against the gain of moving elsewhere after the downtime. In replayed deployments on the Plegma dataset under two kit counts and two downtime costs, ReCo outperforms fixed-dwell and count-based schedules and a threshold rule using the same metric in every setting. Its advantage is not explained by collecting more days alone and is consistent with allocating the days to more valuable homes and periods.
\end{abstract}


\section{Introduction}

Many sensing-based learning tasks acquire their labels by physically deploying instruments at field sites: energy disaggregation, wearable activity recognition and environmental monitoring are typical examples. Such campaigns run under hard resource constraints. Sensor kits are expensive to buy and to install, so for economic reasons a project can deploy only a limited number of them. Access to sites and the project itself are limited in time, so all collection must end by a deadline. Moving a kit is not free either: removal, transport, scheduling and reinstallation take days during which the kit records nothing. Under these constraints, every extra day at one site is a day lost elsewhere: staying too long spends the budget on redundant data, while moving too early pays the downtime repeatedly and may abandon a still-informative site, and this balance shifts as the deadline approaches. The collector must therefore keep deciding from field data whether to stay or move, which no fixed schedule does well for every site.

Non-intrusive load monitoring (NILM)~\cite{hart1992} is a representative instance of this problem. NILM, also called energy disaggregation, estimates how much power each appliance in a home draws using only the household's main meter. This appliance-level breakdown supports energy feedback to residents and demand management without installing a meter on every appliance. Current NILM models are learned from data: a home is temporarily fitted with sub-meters, such as smart plugs or circuit-level meters on the target appliances, which record the true appliance power alongside the aggregate, and a model trained on such homes is then applied to homes that have only the main meter. The sub-meters and the aggregate logger form the sensor kit in our setting, and deciding how long to keep it at each home before relocating it to the next is the collection problem studied here. Two properties of NILM mean that there is no single answer to how long to stay at a home. First, appliance usage frequency varies widely with the appliance, the season and the climate: a refrigerator cycles all day, whereas a washing machine typically runs only a few times per week; an air conditioner may run almost daily in a hot summer yet stay idle for weeks in mild seasons; water-heater use also changes with temperature. The same appliance can therefore yield very different numbers of runs in different homes and periods, and no fixed dwell time suits all cases. Second, the aggregate is the sum of all appliances: a model must learn both the target appliance itself and how it appears on top of different background loads. Whether collected data are useful therefore depends mainly on how diverse the combinations of target operation and background load are.

Existing work has studied related but different questions. ActSense~\cite{jia2019actsense} chooses which home--appliance pairs receive a permanent sub-meter in order to complete a tensor of monthly energy use. Other work considers short-term collection in a single household~\cite{koasidis2023}, or how model performance scales with the amount of data and the number of subjects~\cite{shin2019scale,yong2025,nature2025}. Uncertainty-based criteria have been used to decide when data are sufficient and collection or labeling can stop~\cite{jones2020,sobot2024}; these criteria look only at the data and are not applied under a budget of kits, time and relocation cost. \emph{We focus on the reuse of a fixed number of recoverable kits and on the dwell-time decision under relocation downtime.}

This paper asks: \emph{under constraints on the number of kits, the deadline and the relocation downtime, how should the collector decide from field data when to relocate?} Our contributions are:
\begin{enumerate}
  \item \textbf{A constraint-based relocation framework.} The number of kits, the collection deadline and the cost of relocation downtime are the constraints of the collection framework. Based on these constraints, each night the expected gain of staying one more day is compared with that of relocating after paying the downtime to decide whether to move, so dwell times adapt to what each site yields and to the remaining budget. The framework does not depend on a particular gain measure and could in future be transferred to other field-deployed sensing tasks.
  \item \textbf{A regime-coverage gain for NILM.} As one instantiation of the gain, we count previously unseen operating regimes in a joint space of target-appliance and background features, without training a NILM model.
  \item \textbf{An empirical study.} Under two kit counts and two downtime costs we compare ReCo with fixed dwell, count-based switching and a threshold rule, examine the substitution of an alternative surprise metric and a component ablation of ReCo, and test three diversity scenarios that change the downstream model, the target appliance or the dataset.
\end{enumerate}
The framework is intended for tasks that combine field deployment, synchronous labels and the need to generalize across sites; the coverage metric, however, has to be redesigned and validated for each domain. The empirical conclusions of this paper are limited to NILM data.

\section{Related Work}

\textbf{Sensor deployment for energy disaggregation.} ActSense~\cite{jia2019actsense} applies active tensor completion to monthly energy data to choose which home--appliance pairs receive a permanent sub-meter, so that monthly appliance consumption can be estimated for homes without one. Its meters are never recovered or moved, there is no downtime, and decisions follow model uncertainty; how long a movable kit should stay at a site is not part of the problem.

\textbf{Constrained sensing campaigns.} Koasidis et al.~\cite{koasidis2023} propose an equipment- and time-constrained acquisition protocol that builds a household-specific database for a single home. De Bruin et al.~\cite{sensors2012} decide where and when to move sensors by the expected value of information including movement cost, and their goal is to monitor an environmental state.

\textbf{Scale and allocation.} Studies in speech~\cite{yong2025}, brain imaging~\cite{nature2025} and NILM~\cite{shin2019scale} examine how performance depends on the amount of data and the number of subjects. Vellandurai et al.~\cite{vellandurai2024} use deep reinforcement learning to decide which trips in a transit timetable carry the limited occupancy sensors, so that imputation of the unsensed trips is most accurate. These works do not model relocation downtime or sequential deployment, and the allocation in~\cite{vellandurai2024} is optimized offline for reconstruction accuracy.

\textbf{Novelty- and uncertainty-based stopping.} Criteria based on novelty or uncertainty decide whether to stop, e.g., truncating training data~\cite{jones2020} or stopping label queries~\cite{sobot2024}. They address whether to stop. Our threshold baseline follows this stopping logic, and our surprise-metric substitution turns such a criterion into a stay-or-switch rule.

\textbf{Patch leaving and switching costs.} Our stay-or-switch rule has the form of the marginal value theorem~\cite{charnov1976} and is related to optimal stopping. It also relates to multi-armed bandits with switching costs~\cite{dekel2014}, budget constraints~\cite{badanidiyuru2018} and arms that change while not played~\cite{whittle1988}, which resemble relocation downtime, the collection deadline and seasonal changes in home usage. Bandit policies, however, choose which arm to play in order to learn the best one, whereas here the route is fixed, collecting at a site depletes its novelty, and the only decision is when to leave; adapting bandit policies as baselines is left to future work. Our contribution is its use for constrained field data collection with per-home forecasts of the coverage gain.

\textbf{Active learning for NILM.} Patel et al.~\cite{patel2024} use active learning to choose which home to fit with sub-meters next, and Tanoni et al.~\cite{tanoni2025} combine weak supervision with active learning to choose which data windows users should label. The former decides where to deploy and the latter what to label; neither decides how long a kit should stay at a site.

\textbf{Coverage and data valuation.} Core-set selection~\cite{sener2018}, convex-hull-based data selection (ApproxHull) for NILM~\cite{approxhull}, submodular coverage~\cite{nemhauser1978} and $k$-nearest-neighbor data valuation~\cite{jia2019knn} inspire our regime-coverage metric; RHO-LOSS~\cite{mindermann2022} is a related idea of prioritizing data that are worth learning and not yet learnt.

\section{Problem Formulation}

\subsection{Deployment constraints}
A campaign is characterized by three constraints: (i) $K$ sensor kits, so at most $K$ sites are recorded simultaneously; (ii) a deadline of $T$ days by which all collection must finish; and (iii) a relocation downtime of $c$ days, incurred by a kit every time it \emph{moves} to another site. If a kit makes $n$ visits (a site may be revisited) with dwell times $d_1,\dots,d_n$, then
\begin{equation}
  \sum_{i=1}^{n} d_i + (n-1)\,c \;\le\; T .
  \label{eq:budget}
\end{equation}
The first installation incurs no downtime. For a fair comparison, schedules are compared only under the same $(K,T,c)$; changing $K$ changes the budget and changing $c$ changes the calendar, so absolute values are not comparable across $K$ or $c$, and we never rank schedules across them.

\subsection{Objective}
The collected data train a model of fixed architecture, which is evaluated on sites that took no part in the collection. Formally, given an exogenous visiting route $\pi$, a schedule $\mathcal{S}$ specifies each kit's dwell times and hence its relocation times along $\pi$. With $\mathcal{D}(\mathcal{S})$ the data it collects and $f_{\mathcal{D}(\mathcal{S})}$ the model trained on them, the goal is $\max_{\mathcal{S}\in\Sigma(\pi)}\,\mathbb{E}\big[M(f_{\mathcal{D}(\mathcal{S})})\big]$ for a downstream metric $M$ such as on-state F1, where $\Sigma(\pi)$ is the set of such schedules that satisfy Eq.~(\ref{eq:budget}) for every kit, and the expectation is over the evaluation sites and the training randomness. This objective can only be evaluated after training, so the nightly rule of Section~\ref{sec:method} is a heuristic approximation that uses a data-side gain as a proxy.

\subsection{Replay evaluation protocol}
\begin{figure}[t]
\centering
\begin{tikzpicture}[x=0.235cm,y=0.55cm,font=\scriptsize]
  \draw[-{Latex[length=1.5mm]}] (0,-0.2) -- (30.5,-0.2) node[right]{day};
  \foreach \x in {0,7,14,21,28} \draw (\x,-0.1)--(\x,-0.3) node[below]{\x};
  \node[anchor=east] at (-0.3,1.3) {$K{=}1$};
  \fill[blue!45] (0,1) rectangle (9,1.6);
  \fill[gray!30] (9,1) rectangle (12,1.6);
  \fill[blue!45] (12,1) rectangle (19,1.6);
  \fill[gray!30] (19,1) rectangle (22,1.6);
  \fill[blue!45] (22,1) rectangle (28,1.6);
  \node at (4.5,1.3){home A}; \node at (15.5,1.3){home B}; \node at (25,1.3){home C};
  \node at (10.5,1.3){$c$}; \node at (20.5,1.3){$c$};
  \node[anchor=east] at (-0.3,3.0) {$K{=}2$};
  \fill[blue!45] (0,2.3) rectangle (9,2.9);
  \fill[gray!30] (9,2.3) rectangle (12,2.9);
  \fill[blue!45] (12,2.3) rectangle (28,2.9);
  \node at (4.5,2.6){home A}; \node at (20,2.6){home C}; \node at (10.5,2.6){$c$};
  \fill[orange!55] (0,3.1) rectangle (13,3.7);
  \fill[gray!30] (13,3.1) rectangle (16,3.7);
  \fill[orange!55] (16,3.1) rectangle (28,3.7);
  \node at (6.5,3.4){home B}; \node at (22,3.4){home D}; \node at (14.5,3.4){$c$};
  \draw[dashed,red] (28,-0.2) -- (28,4.0) node[above]{$T$};
\end{tikzpicture}
\caption{Deployment constraints.}
\label{fig:timeline}
\end{figure}
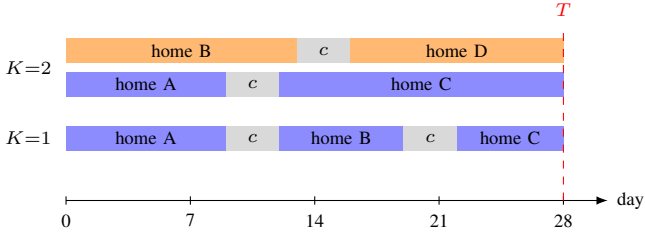

We simulate deployments on a public dataset in which many sites were recorded simultaneously, so that any choice of ``which site, which days'' can be cut from real recordings. All schedules are completed within the same calendar window and differ \emph{only} in when they relocate. Holding the window fixed controls only the coarse time range: the days actually collected still differ between schedules, so date effects cannot be ruled out completely. Fig.~\ref{fig:timeline} illustrates the constraints: each line is one kit, collection alternates with $c$ days of relocation downtime, and all collection ends by the deadline $T$.

\section{Method}\label{sec:method}

We call the full method \textbf{ReCo (Relocation by Coverage gain)}. Each evening it processes only the candidate-home windows collected so far.

Each evening, ReCo asks a simple question: \emph{is one more day at this home likely to show the model something it has not yet seen?} It answers by counting how many genuinely new combinations of target-appliance operation and background load the home has produced so far, and by forecasting how quickly that novelty is running out. This expected gain from staying is then compared with the gain per day of moving on, which includes the days lost to relocation. The kit stays while the current home still pays off better than the next home would after the downtime, and moves once it no longer does. The rest of this section makes each of these three steps precise: counting new regimes (Section~\ref{sec:regime}), forecasting a home's remaining gain (Section~\ref{sec:forecast}), and the stay-or-switch comparison (Section~\ref{sec:rule}).

\begin{figure}[t]
\centering
\begin{tikzpicture}[font=\scriptsize,
  box/.style={draw,rounded corners=2pt,fill=blue!6,align=center,minimum height=0.8cm,text width=1.35cm,inner sep=2pt},
  arr/.style={-{Latex[length=1.5mm]}}]
  \node[box] (w) {collected 2-h windows};
  \node[box,right=0.25cm of w] (n) {new-regime accounting};
  \node[box,right=0.25cm of n] (f) {per-home $\qh,\lh,\ph$};
  \node[box,right=0.25cm of f] (g) {gain forecast $\hat G(h)$};
  \node[box,below=0.45cm of g,fill=orange!12] (d) {stay or switch?};
  \node[box,left=0.25cm of d,fill=gray!10] (r) {default curve from visited-home replay};
  \node[box,left=0.25cm of r,fill=gray!10] (b) {remaining budget $R$, downtime $c$};
  \draw[arr] (w)--(n); \draw[arr] (n)--(f); \draw[arr] (f)--(g); \draw[arr] (g)--(d);
  \draw[arr] (r)--(d); \draw[arr] (b)--(r);
  \draw[arr] (n.south) |- ($(r.north)+(0,0.2)$) -- (r.north);
\end{tikzpicture}
\caption{Pipeline of ReCo.}
\label{fig:pipeline}
\end{figure}
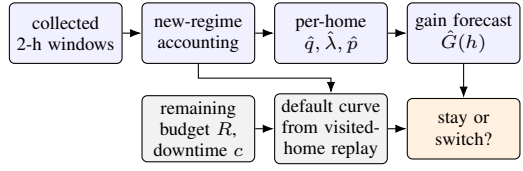

Fig.~\ref{fig:pipeline} summarizes its steps.

\subsection{New-regime accounting}\label{sec:regime}
Each day is divided into twelve two-hour windows, processed in collection order. Windows in which the target appliance is off enter an \emph{off space} described by background power and time-of-day features; such a window is a new regime if no seen off-window lies within a fixed radius. Windows in which the target appliance runs enter a \emph{run space} described by run features and background level (for the washing machine: run duration, energy, peak power, heating duration and background level); such a window is a new regime if no seen window matches it within per-feature tolerances. New regimes join the seen set immediately, so similar windows later on the same day do not score again. The two spaces are counted separately and combined into the daily proxy gain, a measure of collection novelty that is computed without training the model.

\subsection{Per-home forecasts of $q$, $\lambda$ and $p$}\label{sec:forecast}
Each evening ReCo forecasts how many new regimes the current home will still yield. Let $N$ be the windows scheduled at the home, $V$ those that meet the validity rule of Table~\ref{tab:cfg} and have a determinable target state, $A$ the target-active windows among $V$, and $U$ the new-regime windows among $A$; windows whose target state cannot be determined count in $N$ only and are never used as off-windows. ReCo estimates three quantities:
\begin{itemize}
  \item $q$, the share of usable windows, observed as $V/N$;
  \item $\lambda$, the active windows per day if all windows were usable, observed as $12A/V$;
  \item $p$, the probability that an active window is new with respect to the global seen set, observed as $U/A$.
\end{itemize}
Early in a stay these ratios rest on only a few windows and are unstable, so ReCo borrows from the homes it has already seen. Each of the underlying proportions $V/N$, $A/V$ and $U/A$ is estimated from the home's trials $i$ with outcomes $y_i\in\{0,1\}$, which are the scheduled windows for $V/N$, the valid windows for $A/V$ and the active windows for $U/A$. A trial recorded $\Delta t_i$ days ago receives the forgetting weight $w_i=2^{-\Delta t_i/H}$, and
\begin{equation}
  \hat\rho=\frac{\sum_i w_i y_i+1+\kappa m}{\sum_i w_i+2+\kappa},
  \label{eq:shrink}
\end{equation}
shrunk with strength $\kappa$ towards a center $m$, the same proportion estimated from the other homes already visited by that evening; without such homes, $\kappa=0$ and only a weak $\mathrm{Beta}(1,1)$ prior remains. $H=\infty$ gives the unweighted estimate. The more data the current home has, the less it borrows. Each evening $(\kappa,H)$ is chosen from a small candidate set by the one-step-ahead forecast loss, using only the history collected before that evening.

The forecast gain of staying $h$ more days is
\begin{equation}
  \hat G(h) = \sum_{k=1}^{h} \Big[\qh\,\lh\,\ph(E_{k-1}) + \omega\,\qh\,\hat o_k\Big],\qquad E_{k} = E_{k-1} + \qh\,\lh .
  \label{eq:gain}
\end{equation}
The first term is the expected number of new run regimes per day (usable share $\times$ active windows $\times$ novelty rate); the second adds new off regimes, where $\hat o$ is their daily rate under full availability and $\omega$ their weight. The longer ReCo stays, the more of the home's regimes it has already seen, so novelty decays as coverage grows: with $E$ the cumulative number of valid active windows at the home (collected so far plus forecast, updated as in Eq.~(\ref{eq:gain})) and $E_0$ those collected so far, $\ph(E)=p_0\,(E_0+\kappa_p+2)/(E+\kappa_p+2)$, where $p_0$ is tonight's estimate of $p$ and $\kappa_p$ its shrinkage strength. The off rate decays in the same way with off exposure: $\hat o_k=\hat o_1\,(E_{\rm off}+\kappa_{\rm off}+1)/(E_{\rm off}+\kappa_{\rm off}+1+u_k)$ with $u_k=(k-1)\,\qh\,(12-\lh)/12$, where $\hat o_1$ is tonight's estimate, $E_{\rm off}$ the off exposure collected so far and $\kappa_{\rm off}$ its shrinkage strength. The decay applies only to future days; $p_0$ and $\hat o_1$ are re-estimated every evening.

\subsection{Stay or switch}\label{sec:rule}
Each evening ReCo asks whether to spend the next day at the current home or to move. Staying is worth $\hat G_{\text{cur}}(1)$, the gain expected tomorrow; one day suffices because the question is asked again the next evening. Moving first costs $c$ days without data, so the next home is valued by its average gain per day over the downtime plus a stay of $h$ days, using the best $h$ that still fits into the remaining budget $R$:
\begin{equation}
  \text{switch} \iff \max_{h\in\mathcal{H}(R)} \frac{\hat G_{\text{next}}(h)}{c+h} \;>\; \hat G_{\text{cur}}(1),
  \label{eq:rule}
\end{equation}
where $\mathcal{H}(R)$ is the set of such $h$. A longer downtime lowers this average and therefore keeps the kit at a home longer. If the next home was visited before (the route is cyclic), its $\hat G_{\text{next}}$ is forecast from its own history. Otherwise ReCo uses a default curve that describes what a newly visited home typically yields day by day: it replays the first stay of each visited (donor) home in order, counting a window as new if it differs from all valid windows collected up to decision night $t$ at the other visited candidate homes and from the earlier replayed windows of the donor. The donor's own later data, evaluation homes and future data are excluded. Each replayed day scores its new run regimes plus $\omega$ times its new off regimes, so off regimes enter both the stay and the switch side. For a donor whose first stay covers days $1,\dots,L_b$, those days use its leave-one-out replayed gain. From day $L_b{+}1$ on, $\qh$, $\lh$, $\ph$ and $\hat o$ estimated from that prefix are propagated with the per-day exposure recursion of Eq.~(\ref{eq:gain}), without the donor's later revisits or any data after the decision night. Every donor is extended to the largest feasible dwell of the night, the default curve averages the donors with equal weight day by day, and the switch value in Eq.~(\ref{eq:rule}) maximizes the cumulative gain of the first $h$ days divided by $c+h$. A donor observed for 7 days thus contributes replayed gains for days 1--7 and forecasts for days 8--12 when the candidate dwell is 12 days; the dwell is neither truncated to 7 days nor extended by repeating day 7. Each kit starts with a fixed start-up dwell at its first home, whose installation carries no downtime; the minimum dwell and end-of-budget actions are given in Section~\ref{sec:impl}. Algorithm~\ref{alg:reco} summarizes the nightly procedure.

\begin{algorithm}[t]
\caption{ReCo: nightly stay-or-switch decision for one kit}
\label{alg:reco}
\small
\begin{algorithmic}[1]
\Require route $\pi$ (random visiting order); deadline $T$; downtime $c$; start-up dwell $h_0$; minimum dwell $h_{\min}$; candidate shrinkage and forgetting settings $\Theta$
\State install the kit at the first home of $\pi$; $\tau \gets 0$
\For{$t = 1,\dots,T$}
  \State \textbf{if} the kit is in downtime \textbf{then continue}
  \State collect day $t$ at the current home; $\tau \gets \tau+1$
  \State count new off/run regimes against the global seen set and add them to it
  \State \textbf{if} $\tau < h_0$ (first home) \textbf{or} $\tau < h_{\min}$ \textbf{then continue}
  \State $R \gets T - t$
  \State \textbf{if} $R \le c + h_{\min}$ \textbf{or} no free home remains \textbf{then continue}
  \State select $\theta \in \Theta$ by one-step-ahead loss on earlier prefixes
  \State update $\qh,\lh,\ph$ and $\hat o$ for the current home with $\theta$
  \State $G_{\text{stay}} \gets \hat G_{\text{cur}}(1)$ \Comment{Eq.~(\ref{eq:gain})}
  \State build $\hat G_{\text{next}}$ from the next home's own history if revisited, else by time-progressive replay of visited homes
  \State $G_{\text{move}} \gets \max_{h\in\mathcal{H}(R)} \hat G_{\text{next}}(h)/(c+h)$
  \If{$G_{\text{move}} > G_{\text{stay}}$}
    \State move to the next free home on $\pi$; start $c$ days of downtime; $\tau \gets 0$
  \EndIf
\EndFor
\end{algorithmic}
\end{algorithm}

\section{Experiments}

\subsection{Data and evaluation protocol}
We use the washing machine in the Plegma dataset~\cite{plegma2024} (10-s samples, twelve two-hour windows per day). Its eleven usable homes form four folds with two evaluation and nine candidate homes each. The evaluation homes are the eight homes with complete September data, namely \{3,13\}, \{1,7\}, \{6,11\} and \{8,12\}. Collection runs from 1 May to 28 August 2023 ($T{=}120$ days) and evaluation from 1 May to 28 September 2023; collection and evaluation dates may overlap, but homes never do. Evaluation homes are never used for collection, routing, forecasting, tuning or training, and days with missing data still count toward the budget. Each setting is run with three seeds, each drawing a random visiting order that all methods share together with the training setup. Metrics are averaged with equal weight over the two evaluation homes, then over the seeds and finally over the four folds, and the per-fold best baseline is selected on the fold-level values. We report on-state F1, which measures how well the model detects when the washing machine is running (balancing precision and recall; higher is better), and overall MAE, the mean absolute error of the predicted appliance power in watts (lower is better).

\subsection{Features and downstream model}
The regime features are listed in Table~\ref{tab:cfg}. The downstream model is a temporal convolutional disaggregation model (dilated residual TCN with a gated power head, about 3.2M parameters) that maps 721 samples to the central 121 points. All methods share its inputs, training steps, on-threshold and seeds, so only the collection calendar differs.

\subsection{Implementation settings}\label{sec:impl}
Table~\ref{tab:cfg} lists the settings used for the washing machine. The event-rule and run-space values correspond to visible features of the appliance's power waveform: the on-power threshold to the step between standby and running, the gap limit to the longest pause within a wash cycle, and the minimum event length to the shortest complete cycle; the run-space features and tolerances likewise describe how cycles differ from one another. The event and regime settings are fixed before any calendar is generated and are not tuned on the downstream results. We therefore expect that adapting ReCo to another appliance mainly requires reading these values off a few example waveforms of that appliance, although we have not tested this beyond the appliances studied here; a kettle, for instance, would need a higher on-power threshold and no 10-minute minimum, while the stay-or-switch rule itself is unchanged. A run crossing a window boundary counts once for Count-5/10 but enters the regime accounting in each of its windows. For each seed, the nine candidates are visited in a random order, and kits cycle through it. With $K{=}2$, a home hosts at most one kit at a time; kit~1 has priority and the other takes the next free home. When at most $c{+}1$ days remain or no home is free, the kit stays.

\begin{table}[t]
\centering
\caption{Implementation settings of the NILM instance}
\label{tab:cfg}
\footnotesize
\setlength{\tabcolsep}{3pt}
\begin{tabular}{@{}p{0.26\columnwidth}p{0.70\columnwidth}@{}}
\toprule
Item & Setting\\
\midrule
Data and folds & Plegma washing machine; 11 homes, 4 folds; 2 evaluation and 9 candidate homes per fold\\
Time grid & 10-s samples; 12 two-hour windows of 720 samples per day; a window is valid only if $\ge$576 samples have both aggregate and target label and, under the appliance's event rules, gaps change neither its on/off class nor its number of run starts; otherwise it is unknown\\
Budget & $T{=}120$ days; $K\in\{1,2\}$; $c\in\{1,3\}$ days\\
Dwell & start-up dwell $h_0{=}7$ days at the first home of each kit (ReCo, Threshold); minimum dwell $h_{\min}{=}1$ full day for all methods; decisions at the end of a day\\
Target run (washing machine) & label power $>$50\,W; gaps of $\le$110 samples merged; duration $\ge$10\,min\\
Off space & background median power, fluctuation and time of day, scaled by 200\,W, 100\,W and 4\,h (circular time); seen if normalized distance $\le$1.0\\
Run space & same regime if duration $\pm$20\,min, energy $\pm$0.15\,kWh, peak $\pm$300\,W, heating $\pm$10\,min and background $\pm$200\,W all hold\\
Forecast and rule & $\omega{=}0.25$; shrinkage candidates $q$: \{0,12,36,72\} windows, $\lambda$/off: \{0,2,5,10\} valid days (12 valid windows each), $p$: \{0,4,12,24\} active windows; half-life $H\in\{7,14,28,\infty\}$ days; Threshold relocates when the mean gain of the last 3 valid dwell days is $<$1.00 new regimes/day\\
\bottomrule
\end{tabular}
\end{table}

\subsection{Baselines}
Baselines are Fixed-7/14 (relocate after 7 or 14 days), Count-5/10 (relocate after 5 or 10 complete target runs), and a Threshold rule that uses the same regime metric with a fixed threshold of 1.00 new regimes per day. This value was chosen in preliminary runs, in which we tried several values and kept one that behaved reasonably; it was not tuned per setting and is used in all scenarios. The threshold rule averages the gain over the last three valid dwell days and keeps the kit in place until three such days are available. Fixed dwell and count-based switching provide intuitive calendar references; the threshold rule isolates the stay-or-switch rule. With $K{=}2$, the two kits follow the seed's common random candidate route and decide separately every evening, sharing the seen-regime set and home-occupancy state; downtime is charged to each kit's own calendar.

\subsection{Main results}
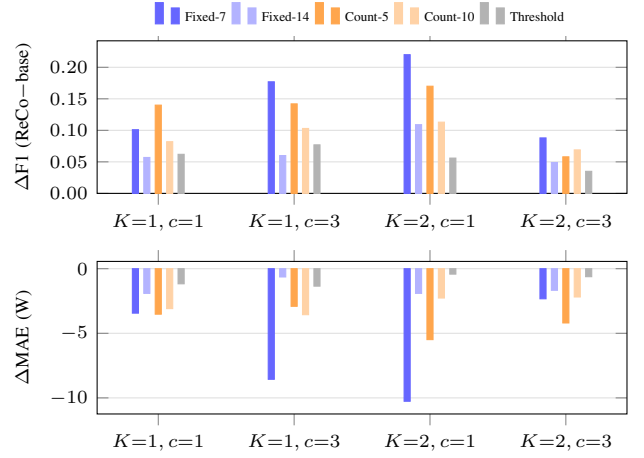
\begin{figure}[t]
\centering
\begin{tikzpicture}
\begin{groupplot}[group style={group size=1 by 2, vertical sep=0.9cm},
  width=0.97\columnwidth, height=3.6cm, ybar, /pgf/bar width=2.3pt,
  symbolic x coords={K1c1,K1c3,K2c1,K2c3},
  xticklabels={{$K{=}1,c{=}1$},{$K{=}1,c{=}3$},{$K{=}2,c{=}1$},{$K{=}2,c{=}3$}},
  xtick=data, tick label style={font=\scriptsize}, label style={font=\scriptsize},
  enlarge x limits=0.15, ymajorgrids, grid style={gray!25},
  legend style={font=\tiny, legend columns=5, at={(0.5,1.03)}, anchor=south, draw=none},
  cycle list={{fill=blue!60,draw=blue!60},{fill=blue!30,draw=blue!30},{fill=orange!70,draw=orange!70},{fill=orange!35,draw=orange!35},{fill=gray!60,draw=gray!60}}]
\nextgroupplot[ylabel={$\Delta$F1 (ReCo$-$base)}, ymin=0, ytick={0,0.05,0.10,0.15,0.20}, scaled y ticks=false, yticklabel style={/pgf/number format/fixed, /pgf/number format/fixed zerofill, /pgf/number format/precision=2}]
\addplot coordinates {(K1c1,0.101) (K1c3,0.177) (K2c1,0.220) (K2c3,0.088)};
\addplot coordinates {(K1c1,0.057) (K1c3,0.060) (K2c1,0.109) (K2c3,0.049)};
\addplot coordinates {(K1c1,0.140) (K1c3,0.142) (K2c1,0.170) (K2c3,0.058)};
\addplot coordinates {(K1c1,0.082) (K1c3,0.103) (K2c1,0.113) (K2c3,0.069)};
\addplot coordinates {(K1c1,0.062) (K1c3,0.077) (K2c1,0.056) (K2c3,0.035)};
\legend{Fixed-7,Fixed-14,Count-5,Count-10,Threshold}
\nextgroupplot[ylabel={$\Delta$MAE (W)}]
\addplot coordinates {(K1c1,-3.44) (K1c3,-8.56) (K2c1,-10.26) (K2c3,-2.33)};
\addplot coordinates {(K1c1,-1.91) (K1c3,-0.64) (K2c1,-1.91) (K2c3,-1.67)};
\addplot coordinates {(K1c1,-3.52) (K1c3,-2.91) (K2c1,-5.49) (K2c3,-4.20)};
\addplot coordinates {(K1c1,-3.09) (K1c3,-3.56) (K2c1,-2.27) (K2c3,-2.19)};
\addplot coordinates {(K1c1,-1.17) (K1c3,-1.35) (K2c1,-0.42) (K2c3,-0.62)};
\end{groupplot}
\end{tikzpicture}
\caption{Paired differences, ReCo minus baseline.}
\label{fig:diff}
\end{figure}

Table~\ref{tab:main} and Fig.~\ref{fig:diff} compare ReCo with each baseline. In all four settings ReCo achieves higher on-state F1 and lower MAE than every baseline. The per-fold best stitching is a stricter reference: in each fold it picks, after the fact, the best of all five baselines including the threshold rule, separately for F1 and for MAE, so the two may come from different methods. ReCo still outperforms it in F1 in all four settings and in MAE in three. The parentheses in the ReCo row of Table~\ref{tab:main} give ReCo minus the per-fold best (mean\,$\pm$\,SD over the four folds). The one exception (MAE at $K{=}1,c{=}3$) is not surprising: the per-fold best is picked after the fact, separately per fold and per metric, and is not available in a real deployment. ReCo adapts to each budget and stays at or near the best baseline chosen in hindsight.

\begin{table*}[t]
\centering
\caption{Main results on Plegma washing machine}
\label{tab:main}
\setlength{\tabcolsep}{4pt}
\providecommand{\abl}[2]{\begin{tabular}[c]{@{}c@{}}#1\\ #2\end{tabular}}
\resizebox{\textwidth}{!}{%
\begin{tabular}{lcccccccc}
\toprule
 & \multicolumn{2}{c}{$K{=}1,\ c{=}1$} & \multicolumn{2}{c}{$K{=}1,\ c{=}3$} & \multicolumn{2}{c}{$K{=}2,\ c{=}1$} & \multicolumn{2}{c}{$K{=}2,\ c{=}3$}\\
\cmidrule(lr){2-3}\cmidrule(lr){4-5}\cmidrule(lr){6-7}\cmidrule(lr){8-9}
Method & F1 & MAE & F1 & MAE & F1 & MAE & F1 & MAE\\
\midrule
Fixed-7   & 0.577 & 11.72 & 0.508 & 19.29 & 0.523 & 17.45 & 0.618 & 10.35\\
Fixed-14  & 0.620 & 10.19 & 0.625 & 11.37 & 0.634 & 9.10  & 0.656 & 9.69\\
Count-5   & 0.537 & 11.81 & 0.543 & 13.64 & 0.573 & 12.68 & 0.647 & 12.22\\
Count-10  & 0.595 & 11.37 & 0.582 & 14.29 & 0.630 & 9.46  & 0.637 & 10.21\\
Threshold & 0.616 & 9.45  & 0.608 & 12.08 & 0.687 & 7.60 & 0.671 & 8.64\\
\midrule
Per-fold best & 0.648 & 8.55 & 0.643 & 10.59 & 0.695 & 7.45 & 0.675 & 8.56\\
\midrule
\textbf{ReCo} & \abl{\textbf{0.678}}{($+$0.030\,$\pm$\,0.013)} & \abl{\textbf{8.29}}{($-$0.26\,$\pm$\,0.04)} & \abl{\textbf{0.685}}{($+$0.042\,$\pm$\,0.007)} & \abl{10.73}{($+$0.14\,$\pm$\,0.10)} & \abl{\textbf{0.743}}{($+$0.048\,$\pm$\,0.006)} & \abl{\textbf{7.19}}{($-$0.26\,$\pm$\,0.04)} & \abl{\textbf{0.706}}{($+$0.031\,$\pm$\,0.015)} & \abl{\textbf{8.02}}{($-$0.54\,$\pm$\,0.06)}\\
\bottomrule
\end{tabular}}
\end{table*}

\subsection{Collection calendars}
Table~\ref{tab:cal} summarizes the calendars: the number of relocations (Sw), the collected days per visit (Dwell) and the valid two-hour windows (Win); collected device-days equal $KT-c\cdot\mathrm{Sw}$. ReCo relocates less often than Fixed-7, Count-5 and the threshold rule. The threshold rule relocates most, staying only 4.6--6.2 days per visit, so downtime consumes the most days and it collects the fewest valid windows. ReCo's advantage is not explained by collecting more days alone: Fixed-14 and Count-10 collect about as many or more device-days and a similar or larger number of windows yet are less accurate. Instead, the results are consistent with ReCo allocating its days to more valuable homes and periods. ReCo also collects the most valid windows per collected day in every setting, so part of its advantage may come from steering kits towards homes and periods with more usable data rather than from regime coverage alone; we do not separate these two effects here.

\begin{table*}[t]
\centering
\caption{Collection calendars}
\label{tab:cal}
\setlength{\tabcolsep}{4pt}
\begin{tabular}{lcccccccccccc}
\toprule
 & \multicolumn{3}{c}{$K{=}1,\ c{=}1$} & \multicolumn{3}{c}{$K{=}1,\ c{=}3$} & \multicolumn{3}{c}{$K{=}2,\ c{=}1$} & \multicolumn{3}{c}{$K{=}2,\ c{=}3$}\\
\cmidrule(lr){2-4}\cmidrule(lr){5-7}\cmidrule(lr){8-10}\cmidrule(lr){11-13}
Method & Sw & Dwell & Win & Sw & Dwell & Win & Sw & Dwell & Win & Sw & Dwell & Win\\
\midrule
\textbf{ReCo} & 9.3 & 10.7 & 1207 & 7.3 & 11.8 & 1083 & 19.3 & 10.4 & 2374 & 11.8 & 14.8 & 2203\\
\midrule
Fixed-7 & 14.0 & 7.1 & 1127 & 11.0 & 7.3 & 923 & 28.0 & 7.1 & 2267 & 22.0 & 7.3 & 1841\\
Fixed-14 & 7.0 & 14.1 & 1196 & 6.0 & 14.6 & 1074 & 14.0 & 14.1 & 2417 & 12.0 & 14.6 & 2130\\
Count-5 & 12.8 & 7.8 & 1124 & 10.8 & 7.4 & 927 & 23.5 & 8.5 & 2267 & 19.8 & 8.3 & 1937\\
Count-10 & 6.3 & 15.6 & 1198 & 5.8 & 15.1 & 1094 & 11.5 & 16.9 & 2431 & 10.5 & 16.7 & 2183\\
Threshold & 20.5 & 4.6 & 1047 & 13.8 & 5.3 & 827 & 37.3 & 5.2 & 2146 & 24.8 & 6.2 & 1773\\
\bottomrule
\end{tabular}
\end{table*}

\subsection{Alternative gain metric}
Because the framework takes the gain measure as an input, the regime-coverage gain can be replaced or improved without changing the stay-or-switch rule. As an alternative we use a surprise metric (\emph{Surprise}), which scores each day by the Bayesian surprise of step events in the aggregate signal. The original work~\cite{jones2020} uses surprise as a stopping criterion; we turn it into a stay-or-switch rule inside the same relocation framework, under the same folds and constraints as ReCo. On each day $j$ the step events in the valid windows update a Bayesian model of step events, and the day's gain $B_j$ is the KL divergence of the updated from the previous model ($B_j{=}0$ for a valid day without steps; a day without any determinable window is missing). With effective-day exposure $e_j=V_j/12$, the current rate is $a_t=\sum_j w_jB_j/\sum_j w_je_j$ with $w_j=2^{-(t-j)/H_B}$, and the forecast for future day $k$ is $\hat g^B_k=\qh\,a_t\,(E_t+\kappa_B+1)/(E_t+\kappa_B+1+(k-1)\,\qh)$, where $E_t$ is the effective-day exposure collected so far. As in ReCo, $H_B$ and $\kappa_B$ are selected from a candidate set by one-step-ahead forecast loss. Staying and switching are compared as in Eq.~(\ref{eq:rule}) with $\hat G^B(h)=\sum_{k=1}^{h}\hat g^B_k$ in place of $\hat G(h)$; the default curve for unvisited homes is rebuilt each night by refitting this step-event model with each visited donor left out, replaying its raw step events and recomputing the KL divergence; days already observed use the replayed values, and only later days are forecast. Table~\ref{tab:b} shows the result, with each entry given as ReCo / Surprise, and their difference (mean\,$\pm$\,SD over folds) below it: ReCo has higher F1 in three settings and lower MAE in all four.

\begin{table}[t]
\centering
\caption{Alternative gain metric}
\label{tab:b}
\setlength{\tabcolsep}{3pt}
\footnotesize
\providecommand{\abl}[2]{\begin{tabular}[c]{@{}c@{}}#1\\ #2\end{tabular}}
\begin{tabular}{lcc}
\toprule
Cell & \abl{F1}{ReCo / Surprise} & \abl{MAE (W)}{ReCo / Surprise}\\
\midrule
$K{=}1,c{=}1$ & \abl{0.678 / 0.639}{($+$0.039\,$\pm$\,0.014)} & \abl{8.29 / 8.94}{($-$0.66\,$\pm$\,0.24)}\\
$K{=}1,c{=}3$ & \abl{0.685 / 0.690}{($-$0.005\,$\pm$\,0.004)} & \abl{10.73 / 13.02}{($-$2.29\,$\pm$\,0.80)}\\
$K{=}2,c{=}1$ & \abl{0.743 / 0.697}{($+$0.046\,$\pm$\,0.007)} & \abl{7.19 / 7.54}{($-$0.35\,$\pm$\,0.07)}\\
$K{=}2,c{=}3$ & \abl{0.706 / 0.668}{($+$0.038\,$\pm$\,0.020)} & \abl{8.02 / 8.57}{($-$0.55\,$\pm$\,0.20)}\\
\bottomrule
\end{tabular}
\end{table}

\subsection{Component ablation}
We compare the full ReCo with three single removals: \emph{w/o availability correction} sets the forecast availability to one, i.e.\ $\qh{=}1$ in Eq.~(\ref{eq:gain}) and in the off exposure $u_k$, and leaves the handling of history unchanged; \emph{w/o home novelty} replaces the per-home novelty rate by a pooled cross-home rate; \emph{w/o adaptive shrinkage/forgetting} fixes the settings to 36 windows for $q$, 5 valid days (60 windows) for $\lambda$ and the off-regime rate, 12 active windows for $p$ and a 28-day half-life; $q/\lambda/p$ are still updated daily, but the settings are no longer reselected by forecast loss.

Table~\ref{tab:abl} shows the results; values in parentheses are the setting minus full ReCo (mean\,$\pm$\,SD over the four folds), so a negative F1 change or a positive MAE change means worse. Removing per-home novelty causes the largest F1 drop in three of four cells ($-$0.090 to $-$0.097) together with MAE increases of 1.57--1.72\,W. Removing the availability correction lowers F1 by 0.057--0.084 in all cells. Its MAE rises in three cells but falls by 0.45\,W at $K{=}1,c{=}3$. Fixing shrinkage and forgetting has the smallest effect on F1 ($-$0.005 to $-$0.053) but consistently raises MAE (0.85--0.97\,W).

\begin{table*}[t]
\centering
\caption{Component ablation}
\label{tab:abl}
\setlength{\tabcolsep}{2.1pt}
\footnotesize
\providecommand{\abl}[2]{\begin{tabular}[c]{@{}c@{}}#1\\ #2\end{tabular}}
\resizebox{\textwidth}{!}{%
\begin{tabular}{lcccccccc}
\toprule
 & \multicolumn{2}{c}{$K{=}1,\ c{=}1$} & \multicolumn{2}{c}{$K{=}1,\ c{=}3$} & \multicolumn{2}{c}{$K{=}2,\ c{=}1$} & \multicolumn{2}{c}{$K{=}2,\ c{=}3$}\\
\cmidrule(lr){2-3}\cmidrule(lr){4-5}\cmidrule(lr){6-7}\cmidrule(lr){8-9}
Setting & F1 & MAE & F1 & MAE & F1 & MAE & F1 & MAE\\
\midrule
Full ReCo & 0.678 & 8.29 & 0.685 & 10.73 & 0.743 & 7.19 & 0.706 & 8.02\\
w/o availability corr. & \abl{0.599}{($-$0.078\,$\pm$\,0.011)} & \abl{9.78}{($+$1.49\,$\pm$\,0.77)} & \abl{0.618}{($-$0.067\,$\pm$\,0.043)} & \abl{10.28}{($-$0.45\,$\pm$\,0.22)} & \abl{0.686}{($-$0.057\,$\pm$\,0.008)} & \abl{8.53}{($+$1.34\,$\pm$\,0.82)} & \abl{0.622}{($-$0.084\,$\pm$\,0.022)} & \abl{9.44}{($+$1.42\,$\pm$\,0.82)}\\
w/o home novelty & \abl{0.588}{($-$0.090\,$\pm$\,0.057)} & \abl{9.94}{($+$1.66\,$\pm$\,0.27)} & \abl{0.589}{($-$0.097\,$\pm$\,0.024)} & \abl{12.46}{($+$1.72\,$\pm$\,1.01)} & \abl{0.647}{($-$0.097\,$\pm$\,0.015)} & \abl{8.75}{($+$1.57\,$\pm$\,0.79)} & \abl{0.635}{($-$0.071\,$\pm$\,0.045)} & \abl{9.69}{($+$1.67\,$\pm$\,0.98)}\\
w/o adapt. shrink./forget. & \abl{0.673}{($-$0.005\,$\pm$\,0.005)} & \abl{9.13}{($+$0.85\,$\pm$\,0.55)} & \abl{0.647}{($-$0.038\,$\pm$\,0.016)} & \abl{11.71}{($+$0.97\,$\pm$\,0.46)} & \abl{0.711}{($-$0.032\,$\pm$\,0.018)} & \abl{8.11}{($+$0.92\,$\pm$\,0.51)} & \abl{0.653}{($-$0.053\,$\pm$\,0.026)} & \abl{8.92}{($+$0.90\,$\pm$\,0.45)}\\
\bottomrule
\end{tabular}}
\end{table*}

\subsection{Diversity scenarios}
Each diversity scenario changes one factor of the main setting and keeps the four constraint settings: the downstream model is replaced by an SGN~\cite{shin2019sgn} adapted to the same input and output (about 74M parameters), the target appliance by the air conditioner, or the dataset by REFIT~\cite{murray2017refit} with the washing machine as target. ReCo, Fixed-14, Count-10 and the threshold rule are compared for $K\in\{1,2\}$. The baseline in Table~\ref{tab:div} is built like the per-fold best above: in each fold, the best of Fixed-14, Count-10 and the threshold rule is taken separately for F1 and for MAE. Because the gain scale differs across appliances and datasets, a fixed threshold may be less well calibrated in these scenarios; Fixed-14 and Count-10, which do not depend on this scale, are therefore also included in the baseline. The air-conditioner scenario uses the Plegma dates, restricted to the homes with air-conditioner channels; all air-conditioner channels of a home are summed as the target (on-power $>$50\,W, complete events $\ge$180\,s, gaps $\le$2100\,s merged). REFIT collects from 4 October 2014 to 31 January 2015 and is evaluated in February 2015, with four folds; its roughly 8-s samples are aligned to the 10-s grid by nearest neighbor (at most 8\,s apart, no interpolation), grid points without a match count as missing, and the washing-machine rules are on-power $>$20\,W, gaps $<$10\,min merged and runs $\ge$10\,min. Results are averaged over three seeds and four folds in every scenario, and the differences in Table~\ref{tab:div} are given as mean\,$\pm$\,SD over the four folds.

\begin{table*}[t]
\centering
\caption{Diversity scenarios}
\label{tab:div}
\setlength{\tabcolsep}{4pt}
\begin{tabular}{llcc}
\toprule
Scenario & Cell & F1: ReCo / baseline ($\Delta\,\pm$\,SD) & MAE (W): ReCo / baseline ($\Delta\,\pm$\,SD)\\
\midrule
\multirow{4}{*}{\shortstack[l]{Model\\$\to$ SGN}}
 & $K{=}1,c{=}1$ & 0.740 / 0.563 (+0.177\,$\pm$\,0.008) & 3.39 / 6.11 ($-$2.72\,$\pm$\,0.56)\\
 & $K{=}1,c{=}3$ & 0.692 / 0.646 (+0.046\,$\pm$\,0.181) & 4.96 / 4.14 (+0.82\,$\pm$\,0.89)\\
 & $K{=}2,c{=}1$ & 0.722 / 0.673 (+0.049\,$\pm$\,0.045) & 3.44 / 4.32 ($-$0.88\,$\pm$\,0.30)\\
 & $K{=}2,c{=}3$ & 0.721 / 0.525 (+0.196\,$\pm$\,0.113) & 3.83 / 5.66 ($-$1.83\,$\pm$\,0.12)\\
\midrule
\multirow{4}{*}{\shortstack[l]{Target\\$\to$ air conditioner}}
 & $K{=}1,c{=}1$ & 0.695 / 0.530 (+0.165\,$\pm$\,0.013) & 34.97 / 47.28 ($-$12.31\,$\pm$\,1.74)\\
 & $K{=}1,c{=}3$ & 0.681 / 0.486 (+0.195\,$\pm$\,0.023) & 38.55 / 53.33 ($-$14.78\,$\pm$\,2.31)\\
 & $K{=}2,c{=}1$ & 0.729 / 0.529 (+0.200\,$\pm$\,0.017) & 32.94 / 42.20 ($-$9.26\,$\pm$\,2.14)\\
 & $K{=}2,c{=}3$ & 0.699 / 0.651 (+0.048\,$\pm$\,0.020) & 36.60 / 44.56 ($-$7.96\,$\pm$\,1.22)\\
\midrule
\multirow{4}{*}{\shortstack[l]{Dataset\\$\to$ REFIT}}
 & $K{=}1,c{=}1$ & 0.595 / 0.639 ($-$0.044\,$\pm$\,0.058) & 7.66 / 9.57 ($-$1.91\,$\pm$\,0.60)\\
 & $K{=}1,c{=}3$ & 0.655 / 0.488 (+0.167\,$\pm$\,0.028) & 7.97 / 10.12 ($-$2.15\,$\pm$\,0.35)\\
 & $K{=}2,c{=}1$ & 0.679 / 0.497 (+0.182\,$\pm$\,0.039) & 7.16 / 9.58 ($-$2.42\,$\pm$\,0.49)\\
 & $K{=}2,c{=}3$ & 0.601 / 0.646 ($-$0.045\,$\pm$\,0.004) & 7.50 / 10.20 ($-$2.70\,$\pm$\,0.61)\\
\bottomrule
\end{tabular}
\end{table*}

Table~\ref{tab:div} compares ReCo with this per-fold best baseline. ReCo largely keeps its advantage when the model, the target appliance or the dataset changes: ReCo's F1 exceeds the strongest baseline in ten of twelve combinations (by more than 0.05 in seven). In the remaining two, both on REFIT, it is lower by about 0.045 while its MAE remains lower. Its MAE is higher in one case (SGN, $K{=}1,c{=}3$). The SGN reaches a much lower MAE than the main TCN (e.g., 3.39 vs.\ 8.29\,W for ReCo at $K{=}1,c{=}1$); we attribute this mainly to its much larger capacity (about 74M vs.\ 3.2M parameters), while the schedules are always compared within the same model.

\section{Discussion}

\emph{Why coverage and the decision rule matter.} Window counts alone do not explain the differences between methods. The threshold rule collects the fewest valid windows, yet its F1 exceeds that of Fixed-7 and Count-5 in every setting, and at $K{=}2$ it is the best single baseline in both F1 and MAE. Both it and ReCo leave a home when its new regimes run out, which suggests that the regime-coverage metric itself accounts for part of the gain over fixed-dwell and count-based schedules. The effect is not uniform, however: at $K{=}1$ Fixed-14 is slightly ahead of the threshold rule in F1, and at $K{=}1,c{=}3$ also in MAE, so a coverage signal alone does not guarantee a better calendar. The decision rule adds a further improvement on top of the coverage metric, whose size depends on the setting and the metric: the threshold rule leaves as soon as recent gain drops, and part of its budget goes to downtime that could have yielded further regimes. Since ReCo differs from the threshold rule in both its per-home forecasts and its cost-aware comparison, this improvement reflects both rather than the downtime term alone. ReCo weighs this cost and improves both F1 and MAE in all four settings, although the margin varies and is smallest in F1 at $K{=}2,c{=}3$, so the benefit of the decision rule is not uniform.

\emph{What the ablations suggest.} Modeling novelty per home gave the most consistent benefit. Homes differ in how quickly they run out of new regimes, so a rate pooled across homes tends to keep a kit too long at a repetitive home and to move it too early from a varied one. Accounting for data availability also mattered: without it, the gain obtainable under valid observation is taken as the gain actually obtainable over calendar time, which distorts the comparison between staying and switching. Reselecting shrinkage and forgetting online had a smaller effect, mainly on MAE, which suggests that the forecasts are reasonably robust to these settings but still benefit from adapting to each home.

\emph{Future directions.} Because the stay-or-switch rule takes the gain measure as an input, the most direct extension is a better gain: one that is weighted towards the final metric, that uses model uncertainty or expected error reduction once a first model is available, or that is learned from past campaigns. A second direction is to decide where to go next as well as when to leave, combining ReCo with active selection of the next home instead of a random route. Coordinating several kits more closely than by simple priority, handling several target appliances at once, and including monetary cost and travel explicitly in the downtime are further steps towards real campaigns. Finally, a field deployment would test the rule under real installation, access and failure conditions.

\section{Limitations}

Our evaluation replays recorded data instead of deploying kits in the field, so practical factors such as installation, site access and equipment failures are not yet reflected. The evidence is also limited in size: the main study uses eleven homes, three seeds and four folds whose candidate pools overlap, so the fold-level standard deviations give only limited evidence of robustness to other campaigns, seasons and routes, and we do not report significance tests. The baselines cover only two fixed dwell times and two count thresholds, the threshold rule uses one value in all scenarios, and no random-relocation baseline is included. Because ReCo differs from the threshold rule in both its forecasts and its downtime-aware comparison, we have not isolated the contribution of the downtime term, nor tested the effect of the deadline separately. We have not analyzed sensitivity to $\omega$, the regime tolerances or the shrinkage candidate sets, reported forecast accuracy, or directly measured which homes and periods each schedule spends its days on. The main study covers one appliance in one country, and on REFIT a lower MAE did not always come with a higher F1; moreover, overall MAE is dominated by off periods for a rarely used appliance, so event-level errors may be more informative. The multiplicative forecast is also a simplification: it does not model dependence between its factors and is less reliable on a first visit with little history. Finally, this work currently focuses on NILM; applying the framework to other sensing-based collection tasks remains to be explored. We plan to address these points in an extended version.

\section{Conclusion}

We studied data collection for sensing-based learning when kits are few, time is limited and every relocation costs downtime, and framed it as a nightly decision between staying at the current site and moving on. The resulting framework compares the gain of staying with the gain achievable elsewhere after the downtime and treats the gain measure as a replaceable input. For NILM we instantiated it as ReCo, which counts new operating regimes of the target appliance against its background and forecasts, for each home, how many more it is likely to yield.

In replayed deployments on real household data, ReCo outperformed fixed-dwell and count-based schedules under every combination of kit count and downtime, and stayed at or near the best baseline chosen in hindsight. These results are consistent with allocating the collection days to more valuable homes and periods, and ReCo's advantage largely carried over when the downstream model, the target appliance or the dataset was changed. Overall, they suggest that treating relocation as an explicit, budget-aware decision is a simple and effective way to spend a limited collection budget, and that the framework offers a natural place to plug in better gain measures in future work.

\end{document}